\documentclass[letterpaper, 10 pt, conference]{ieeeconf}  

\IEEEoverridecommandlockouts                              

\usepackage{graphics} 
\usepackage{epsfig} 
\usepackage{mathptmx} 
\usepackage{times} 
\usepackage{amsmath} 
\usepackage{amssymb}  
\usepackage{bm}
\usepackage{biblatex}
\usepackage{xcolor}
\usepackage{graphicx}
\graphicspath{ {./images/} }
\usepackage{subcaption}
\usepackage[hidelinks]{hyperref}
\usepackage{sidecap}

\usepackage{caption}
\title{\LARGE \bf
NeuroSMPC: A Neural Network guided Sampling Based MPC for On-Road Autonomous Driving
}

\author{Kaustab Pal$^{1}$, Aditya Sharma$^{1}$, Mohd Omama$^{1}$, Parth N. Shah$^{1}$, K. Madhava Krishna$^{1}$
\thanks{$^{1}$Robotics Research Center, IIIT-Hyderabad, India}
}

\begin{document}

\maketitle
\thispagestyle{empty}
\pagestyle{empty}

\begin{abstract}

In this paper we show an effective means of integrating data driven frameworks to sampling based optimal control to vastly reduce the compute time for easy adoption and adaptation to real time applications such as on-road autonomous driving in the presence of dynamic actors. Presented with training examples, a spatio-temporal CNN learns to predict the optimal mean control over a finite horizon that precludes further resampling, an iterative process that makes sampling based optimal control formulations difficult to adopt in real time settings. Generating control samples around the network-predicted optimal mean retains the advantage of sample diversity while enabling real time rollout of trajectories that avoids multiple dynamic obstacles in an on-road navigation setting. Further the 3D CNN architecture implicitly learns the future trajectories of the dynamic agents in the scene resulting in successful collision free navigation despite no explicit future trajectory prediction. We show performance gain over multiple baselines in a number of on-road scenes through closed loop simulations in CARLA. We also showcase the real world applicability of our system by running it on our custom Autonomous Driving Platform (AutoDP).
\end{abstract}

\section{Introduction}\label{Intro}

On road autonomous driving entails persistent and consistent real-time decision making in often highly dynamic and evolving scenes. To accomplish this most trajectory planning frameworks employ a scheme of generating multiple candidate trajectory proposals and scoring them according to an appropriate cost function \cite{MP3,zeng2019end,CVAE}. The candidate trajectories are typically obtained from large scale driving data \cite{MP3} or by sampling from a parameter distribution that defines a trajectory \cite{zeng2019end}

As an alternative sampling based optimization paradigms have been popular in the high dimensional planning literature that seamlessly integrate dynamical model of the systems over which the trajectory plans are computed. The major concern however here is the disadvantage in rolling out trajectories in real-time

In this paper we propose a novel framework that interleaves neural network driven prediction of finite horizon controls with samples drawn from the variance centred around the mean predicted by the network. The deep network
outputs a vector of controls that constitute the optimal mean control over a finite horizon. The network is supervised with the best controls from the elite samples of the sampling based control framework \cite{pmlr-v164-bhardwaj22a}. Typically this mean is expected to be close to the global optimum \cite{bharadhwaj2020model, CEM-Efficient}. The best rollout sample is one that optimizes a scoring or cost function detailed later. 

\begin{figure}
    \centering
    \begin{subfigure}[t]{\linewidth}
        \includegraphics[width=\textwidth]{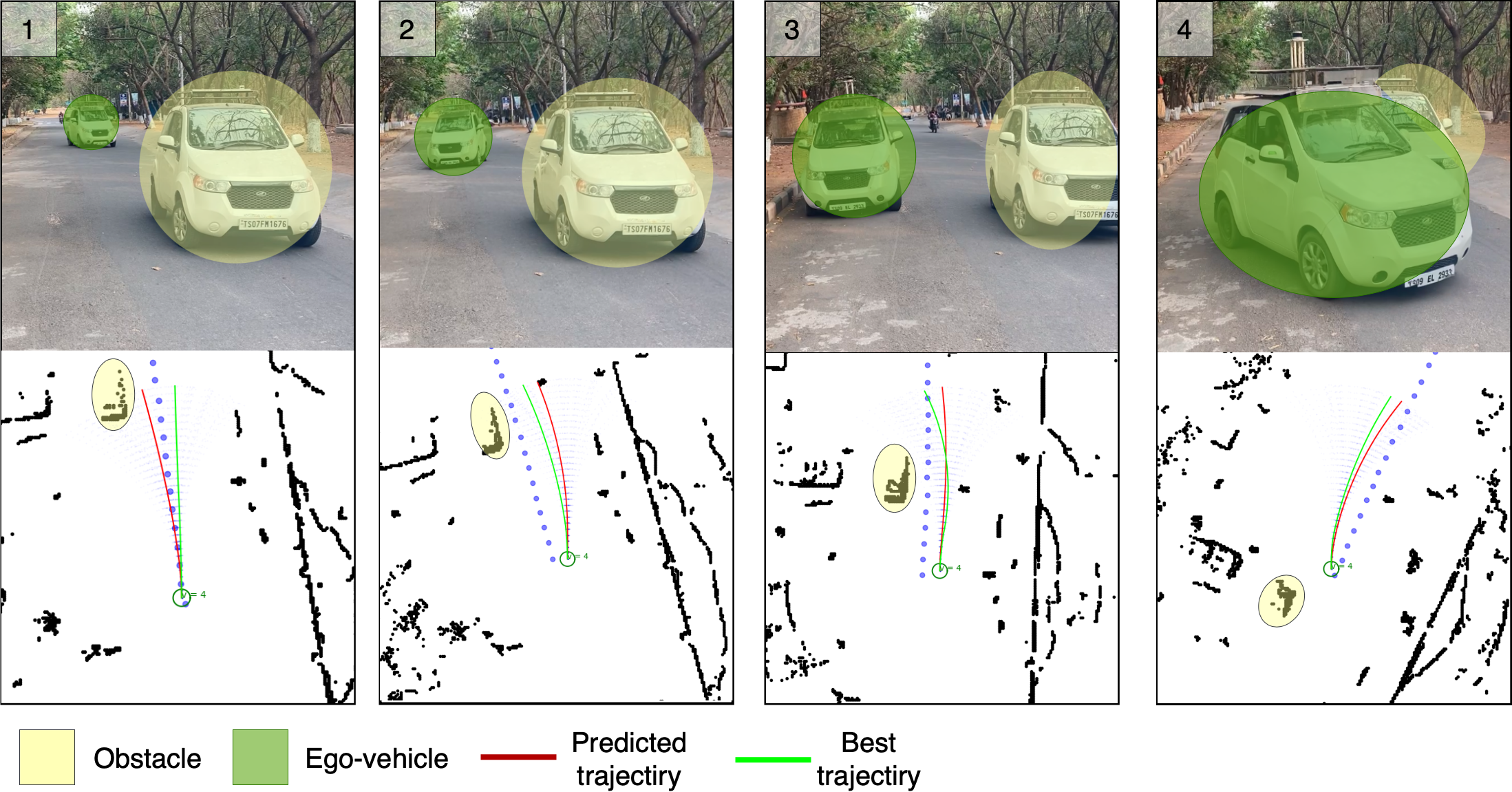}
    \end{subfigure}
    \caption{\textbf{Run on a real car:} We show that our model trained on synthetic data from a simulation is able to run successfully in a real world setting. We ran the model on a real self driving car that avoids an obstacle in front of it in a controlled on-campus scene to ensure safety. The sequences of images in the top row shows that the ego-vehicle (highlighted in green) is able to avoid the obstacle (highlighted in yellow). The sequences of images in the bottom row shows the corresponding occupancy grid maps along with it's predicted trajectory (red) and the best sampled trajectory (green) in the ego-vehicle coordinate frame. }
    \label{fig:sim2real}
\end{figure}

We retain the advantages accrued due to the original framework by not executing the network output controls but choosing the best rollout candidate from the samples around the network output controls. At the same time we overcame the curse of time complexity of such frameworks by selecting the optimal mean obtained from the network that is at-least three orders faster. It is to be noted that almost all such formulations involve many iterations of the following two steps to converge to the optimal control :
\begin{itemize}
    \item Selection of a subset of the candidate samples (the elite set) based on a scoring function. 
    \item Update of the mean in accordance with the elite samples and re-sampling from the newly updated distribution.
\end{itemize}
Apart from this, the network through its spatio-temporal convolution (3D-CNN) architecture implicitly learns the evolution of dynamic actions over time without resorting to explicit trajectory prediction. Despite the lack of explicit trajectory prediction the executed trajectory is collision free in diverse set of on-road navigation scenarios with many dynamic actors. We attribute this to extremely low latency, high frequency control roll-outs possible through the deep network.

The paper contributes specifically in the following ways:
\begin{enumerate}
    \item A neural network interleaved sampling based optimal control that computes finite horizon control rollout in real-time thereby making it suitable for real-time self driving for on-road scenes.
    \item We show 3D convolutions that convolve spatial and temporal components of a time sequenced Birds Eye View (BEV) layout learns implicitly the future trajectories of the dynamic obstacles so much so the planner based on the network output avoids collision with dynamic obstacles despite lack of explicit trajectory prediction into the future.
    \item A number of closed loop simulations in diverse scenarios in the CARLA simulator as well as real-time on campus navigation through our AutoDP confirm the efficacy and real-time veracity of the proposed framework.
    \item Moreover the comparative analysis on compute time with other sampling based optimal control frameworks \cite{pmlr-v164-bhardwaj22a, bharadhwaj2020model} clearly depicts the vast performance gain of the proposed method.
    \item Further, we propose \textit{RoadSeg}, a system for on board real-time road segmentation in known environments that can efficiently generate BEVs in resource constrained setups.
\end{enumerate}

\section{Related Work}\label{RW}

A large portion of on-road driving literature is devoted to getting layout representations \cite{WACV-20, Kendall, LSS}, agent trajectory prediction \cite{Desire, INFER} and end to end trajectory generation \cite{LSS, zeng2019end}. Most of the trajectory generation frameworks involve choosing the best possible trajectory out of an elite set that is typically obtained from large dataset of driving examples or expert trajectories \cite{zeng2019end, LSS}. However the expert trajectories despite the diversity can be sub-optimal and need not be the best response for a given input scenario. Most of these methods do not show their trajectory planning in closed loop simulation scenarios as they are evaluated in pre-recorded datasets that prevent perception and planning in coupled closed loop settings. Also none of the above methods showcase their formulation on a real self driving vehicle in real world on-road scenes.

Elsewhere, in the manipulation planning community, sampling based optimal control \cite{CEM, MPPI} have become popular essentially due to the derivative free blackbox optimization feature of such frameworks. Such sampling based trajectory optimization methods suffer from the curse of time complexity with a number of variants being proposed to overcome this disadvantage. For instance \cite{pmlr-v164-bhardwaj22a} resorts to efficient parallelization of controls while \cite{CEM-Efficient} contributes through sample efficient methods. Whereas in \cite{bharadhwaj2020model} gradient based updates of sample parameters leads to enhanced performance. Nonetheless these methods still need iterative improvement of the sample parameters to reach optimum mean control values that preclude their adaptation to on-road real time autonomous driving applications. 

In contrast to above methods we showcase real time closed loop simulations on a number of CARLA scenes along with a closed loop implementation on our AutoDP to achieve point to point on campus autonomous driving. Moreover our low latency control rollout facilitates navigation in dynamic scenes without collisions despite the lack of explicit trajectory prediction into the future of the dynamic actors.

\section{Methodology}

\begin{figure*}[t]
    \centering
    \begin{subfigure}{\textwidth}
        \includegraphics[width=\textwidth]{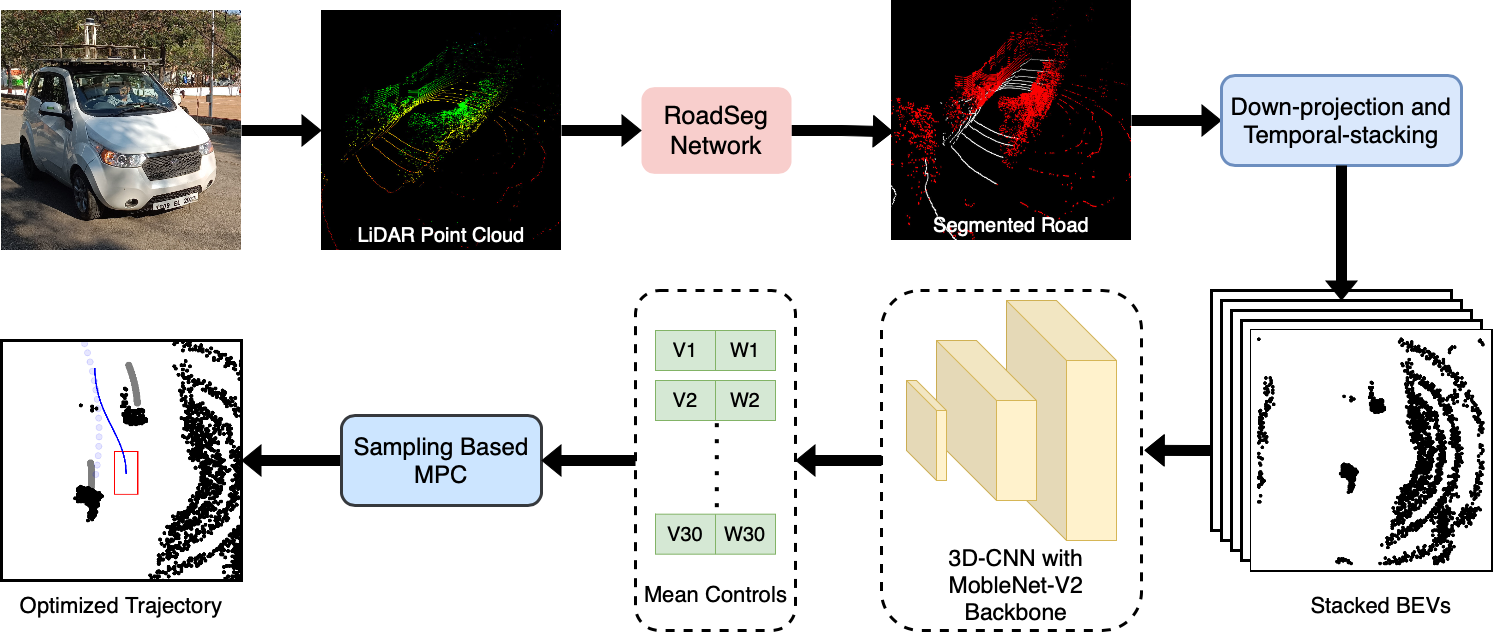}
    \end{subfigure}
    \caption{\textbf{Pipeline:} Each Point-cloud from our AutoDP is first passed through the RoadSeg Network to segment the points that belong to the road. The non-road points are considered as obstacles. A down-projection operation is performed to the obstacle points to create a BEV occupancy grid map. $5$ past BEVs are then stacked together and passed through a 3D-CNN architecture to get the mean controls for a short-horizon. These mean controls are then used as the mean of a sampling based MPC to sample more trajectories. The best trajectory is chosen from amongst these trajectories to be executed by our AutoDP. }
    \label{fig:pipeline}
\end{figure*}

\subsection{Problem statement}
For an on road driving scene, given a point cloud and the global path that our AutoDP needs to follow, we consider the problem of generating occupancy grid maps from the point cloud data and use the occupancy grid map to generate controls that drives the AutoDP towards obstacle free regions on the road while moving along its global path. 
 The state of the vehicle at timestep $h$ is represented by the vector $ \bm{\widetilde{x}_h} = \begin{bmatrix}x_h, y_h, \theta_h\end{bmatrix}$ where $x_h$ and $y_h$ represents the $x$ and the $y$ coordinate of the vehicle respectively and $\theta_h$ is the orientation of the vehicle. The controls for the vehicle at timestep $h$ are represented by the vector $\bm{u_h} = \begin{bmatrix}v_h, \omega_h \end{bmatrix}$ where $v_h$ and $\omega_h$ are the velocity and the angular-velocity of the vehicle respectively.

\subsection{Pipeline}

We show the pipeline of our proposed framework in Fig. \ref{fig:pipeline}. The point-cloud from our AutoDP is first passed through the RoadSeg Network to segment the points that belong to the road. The points that do not belong to the road are considered as obstacles. A down-projection operation is performed to the obstacle points to create a birds-eye view (BEV) occupancy grid map. $5$ of these BEVs are then stacked together and passed through a 3D-CNN CNN based architecture to get the mean controls. These mean controls are used as the mean of a sampling based MPC to sample more trajectories. The best trajectory is chosen from amongst these trajectories to be executed by our AutoDP.

\subsubsection{\textbf{RoadSeg Network}}

\begin{figure}
    \centering
    \includegraphics[width=\linewidth]{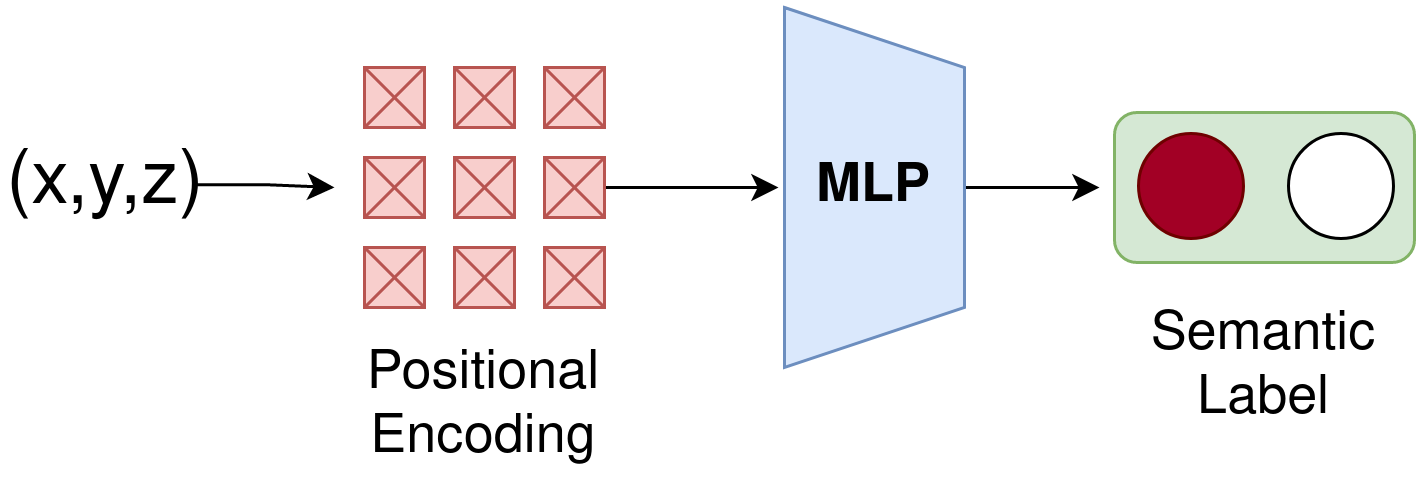}
    \caption{The RoadSeg network takes in the point postions as input, passes them thorough a nerf-like positional encoding followed by a fully connected layer. The result is a segmentation score (road/not-road) for each point.}
    \label{fig:roadseg}
\end{figure}

For the system to work in real time on a resource constrained setup, its essential that the individual building blocks have minimal GPU footprint and the highest possible FPS. The major bottleneck in our pipeline is road-segmentation, which is usually a GPU-heavy process. We can take advantage of the fact that our AutoDP is meant to operate in a known map. We can pre-segment the road using heavy neural networks, then distill this information with a smaller network that can work in real time. The resultant network is like an implicit representation of the operational area wherein the robot can navigate \cite{conceptfusion}. To do this, we first create a map of the operational area of our vehicle using LEGO-LOAM \cite{shan2018lego}. This map is used for global localization and global planning, similar to other autonomous driving platforms \cite{autoware}. The map is created with our custom calibrated-lidar-camera setup. Since, we have 3D points and corresponding RGB values with the calibrated setup, we can use off the shelf segmentation models to segment out the road in the 3D map offline. The choice of the segmentation model is arbitrary.  Once, the road points are segmented out, we can distill this information with a smaller neural network, which we call the RoadSeg network. This network takes in the $x,y,z$ points which are passed to a NERF-like positional encoder \cite{mildenhall2021nerf} followed by few MLP layers. The network's job is to predict if the current points belong to the road or not and is trained using the offline-segmented LEGO-LOAM map as ground-truth. Though the LEGO-LOAM map is very sparse, the positional encoding allows the network to learn general spatial trends and makes it capable of segmenting out the road of the entire (much denser) lidar point clouds at inference. Its also capable of segmenting out regions that were previously not seen in the map creation process (like those occupied by other vehicles). The RoadSeg network is shown in Fig \ref{fig:roadseg}. We show the advantage of using our RoadSeg Network over other baselines in Sec. \ref{quant_analusis}


\subsubsection{\textbf{3D-CNN based Neural Network}}
The occupancy grid map at the current timestep $T$ along with the past $4$ occupancy grid maps and the global path are concatenated into a $6 \times H \times W$ spatio-temporal tensor. This spatio-temporal tensor is then passed into our 3D-CNN based encoder architecture to extract a feature vector. The feature vector is then passed through $4$ fully-connected layers to get an output vector of dimension $H \times 2$ which represents the optimal controls (velocity and angular-velocity). The feature vector from the encoder architecture encodes the spatio-temporal correlations between the occupancy grid maps. For our implementation we have used a slightly modified verson of MobileNet-V2 \cite{sandler2018mobilenetv2} as the encoder.

\subsubsection{\textbf{Sampling based MPC}} \label{smpc}
The objective is to generate controls for short horizons of $H$ timesteps into the future. The output of the 3D-CNN based Neural Network is used as a mean for a Gaussian distribution. We now sample $N$ control sequences of length $H$ from this Gaussian distribution and roll them out using the unicycle kinematics model of the vehicle to get $N$ trajectories. Each of these trajectories are then scored with two cost functions:
\begin{itemize}
    \item \emph{Smoothness cost}: The smoothness cost ensures that we always give more preference to a trajectory with a smooth change in it's linear and angular velocities. 
    \begin{align}
    \hat{c}_{ang}(\bm{\hat{x}_i}, \bm{u_i}) = \sqrt{\sum_{h=1}^{H-1}(u_{w_h} - u_{w_{h-1}})^2} \\
    \hat{c}_{lin}(\bm{\hat{x}_i}, \bm{u_i}) = \sqrt{\sum_{h=1}^{H-1}(u_{v_h} - u_{w_{v-1}})^2}
    \end{align}
    
     where $u_{v}$ and $u_{w}$ represents the linear and angular velocity components of the sampled controls respectively.
    
    \item \emph{Obstacle avoidance cost}: The obstacle avoidance cost ensures that the trajectories are not colliding with any obstacles (occupied cells in the occupancy grid map). 

    \begin{align}
    d(\bm{\widetilde{x}_h}, \bm{\widetilde{o}_h}) = \mid \mid \bm{\widetilde{x}_h} - \bm{\widetilde{o}_h} \mid \mid _2
    \end{align}
    
    Here $\bm{\widetilde{x}_h}$ and $\bm{\widetilde{o}_h}$ are the states of the agent and the obstacle respectively at time-step $h$. The agent and the obstacles are represented as circles. Let $r_A$ and $r_O$ be the radius of the agent and the obstacle respectively. The state $\bm{\widetilde{x}_h}$ is said to be in collision with the obstacle state $\bm{\widetilde{o}_h}$ if the euclidean distance $d(\bm{\widetilde{x}_h}, \bm{\widetilde{o}_h})$ between the agent's position and the obstacle's position is less than or equal to the sum of their radius.
    
    
    \begin{align}
    \hat{c}_{obs}(\bm{\hat{x}_i}, \bm{u_i}) = 
    \begin{cases}
    \infty \text{ , if $d(\bm{\widetilde{x}_h}, \bm{\widetilde{o}_h}) \leq r_A + r_O  $, $h\in[0,H)$}\\
    0 \text{ , otherwise}
    \end{cases} 
    \end{align}
    where $\bm{\hat{x_i}}$ and $\bm{\hat{u_i}}$ are the i-th trajectory and controls out of the $N$ sampled trajectories and controls.

\end{itemize}

\begin{figure}
    \centering
    \includegraphics[width=\linewidth]{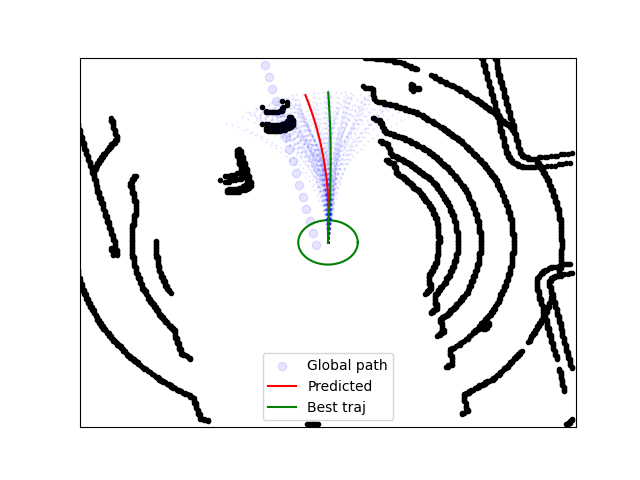}
    \caption{This figure demonstrates the output trajectory (red) of the neural network, the sampled trajectories (blue) from a Gaussian distribution with the neural network output as the mean, and the best trajectory (green) selected from the sampled trajectories after scoring them with the smoothness and obstacle avoidance cost functions.}
    \label{fig:traj_select}
\end{figure}

The final cost $\hat{C}(\bm{\hat{x}_i}, \bm{u_i})$ for each trajectory is calculated as the weighted sum of the three costs

\begin{align}
\hat{C}(\bm{\hat{x}_i}, \bm{u_i}) &= w_{ang} \hat{c}_{ang}(\bm{\hat{x}_i}, \bm{u_i}) + w_{lin} \hat{c}_{lin}(\bm{\hat{x}_i}, \bm{u_i}) + 
\notag \\
& w_{o} \hat{c}_{obs}(\bm{\hat{x}_i}, \bm{u_i})
\end{align}

where $w_{ang}$, $w_{lin}$ and $w_{o}$ are the weights chosen by the user. The trajectory with the smallest cost is chosen as the best trajectory and the controls at $h = 1$ are used to drive the ego-vehicle after which we recompute a new trajectory again. Figure \ref{fig:traj_select} shows the output of the neural network and the best trajectory.

\subsection{Dataset}




Each sample in our dataset consists of a sequence of $5$ occupancy-grid maps along with the global path as the input and the short horizon optimal control as the output. We used the CARLA simulator \cite{Dosovitskiy17} to create a dataset. Obstacles were spawned with varying velocities randomly amongst the CARLA maps. Each occupied cell in the occupancy grid map is used as an obstacle while planning the optimal controls using the Sampling based MPC (SMPC). Since our SMPC operates in the center line reference frame (CRF), we transform all the obstacles to the frame attached to the global path\cite{werling2010optimal}. This allows us to treat curved roads as straight roads and plan feasible trajectories easily without taking the curvature bound constraints. To plan the trajectories that avoids collision with dynamic obstacles whose velocities are known from the CARLA ground truth data, we used the SMPC with the MPPI update rule. The optimal trajectories are then transformed back to the global frame so that the trajectories are within the curvature bounds of the roads in the global frame. 

During inference the BEV doesnot need to be converted from the global frame to the CRF which is a big advantage of our method. Also the 3D-CNN layers learns the dynamic nature of the scene implicitly so we also don't need a separate obstacle trajectory predictor. 

\section{Experimental Evaluation}

\subsection{Qualitative Analysis}

\begin{figure}[t]
    \centering
    \begin{subfigure}{\linewidth}
      \begin{subfigure}{0.44\linewidth}
        \includegraphics[width=\linewidth]{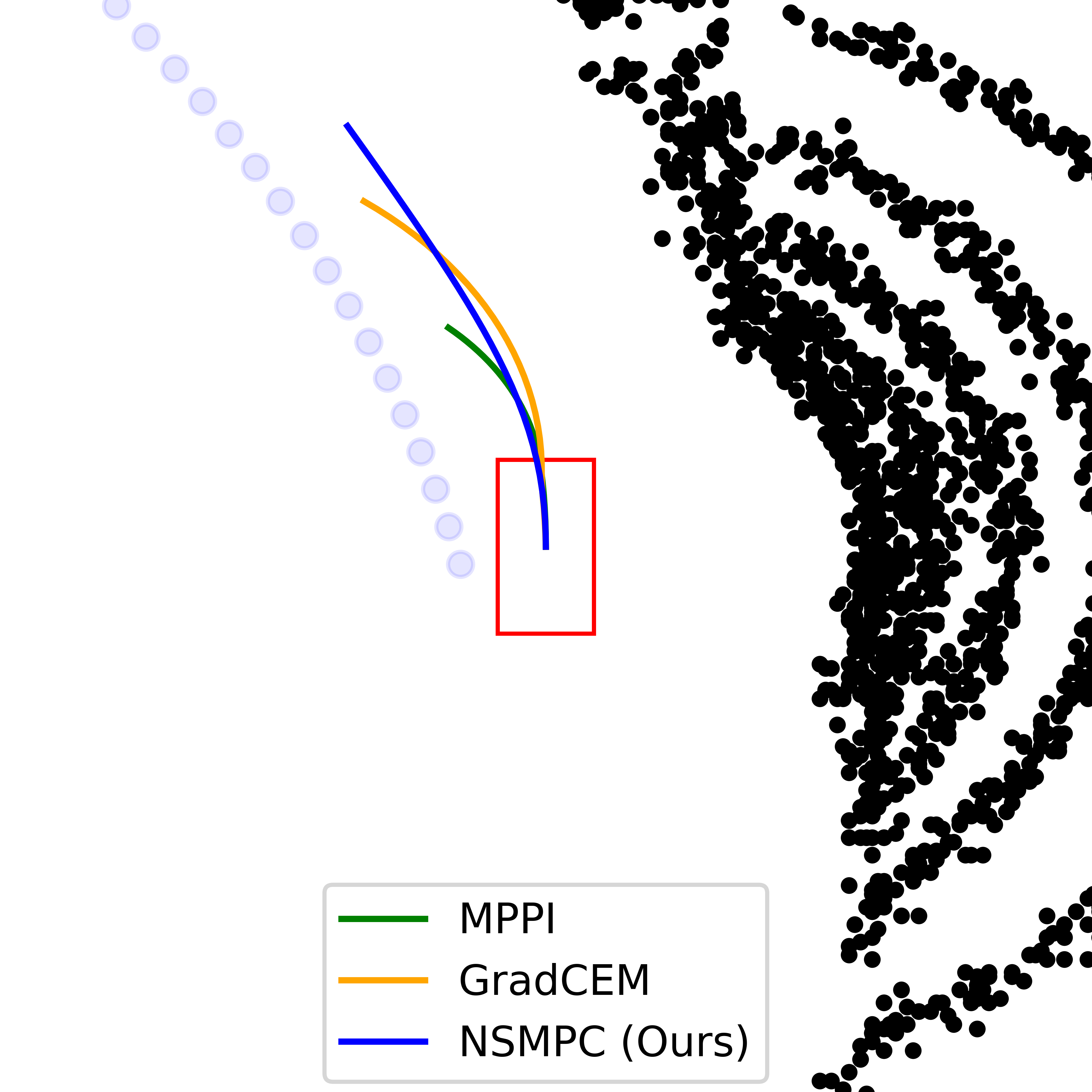}
      \end{subfigure}
      \begin{subfigure}{0.44\linewidth}
        \includegraphics[width=\linewidth]{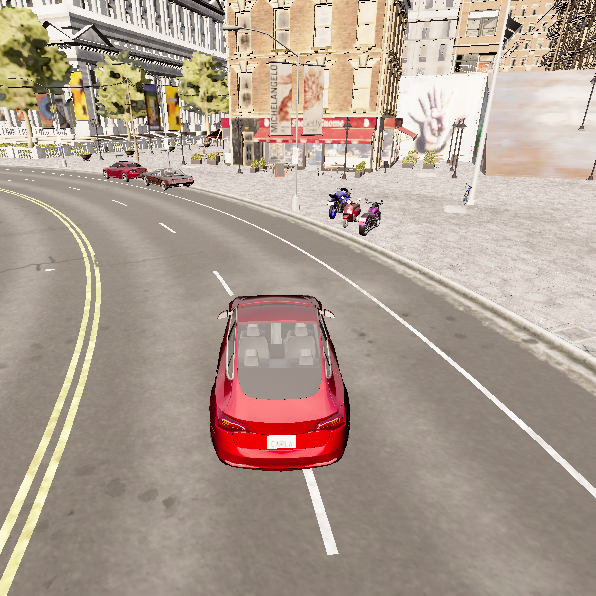}
      \end{subfigure}
      \caption{Curved road} \label{fig:qa_st_lane_scene}
    \end{subfigure}
      \medskip
    \begin{subfigure}{\linewidth}
      \begin{subfigure}{0.44\linewidth}
        \includegraphics[width=\linewidth]{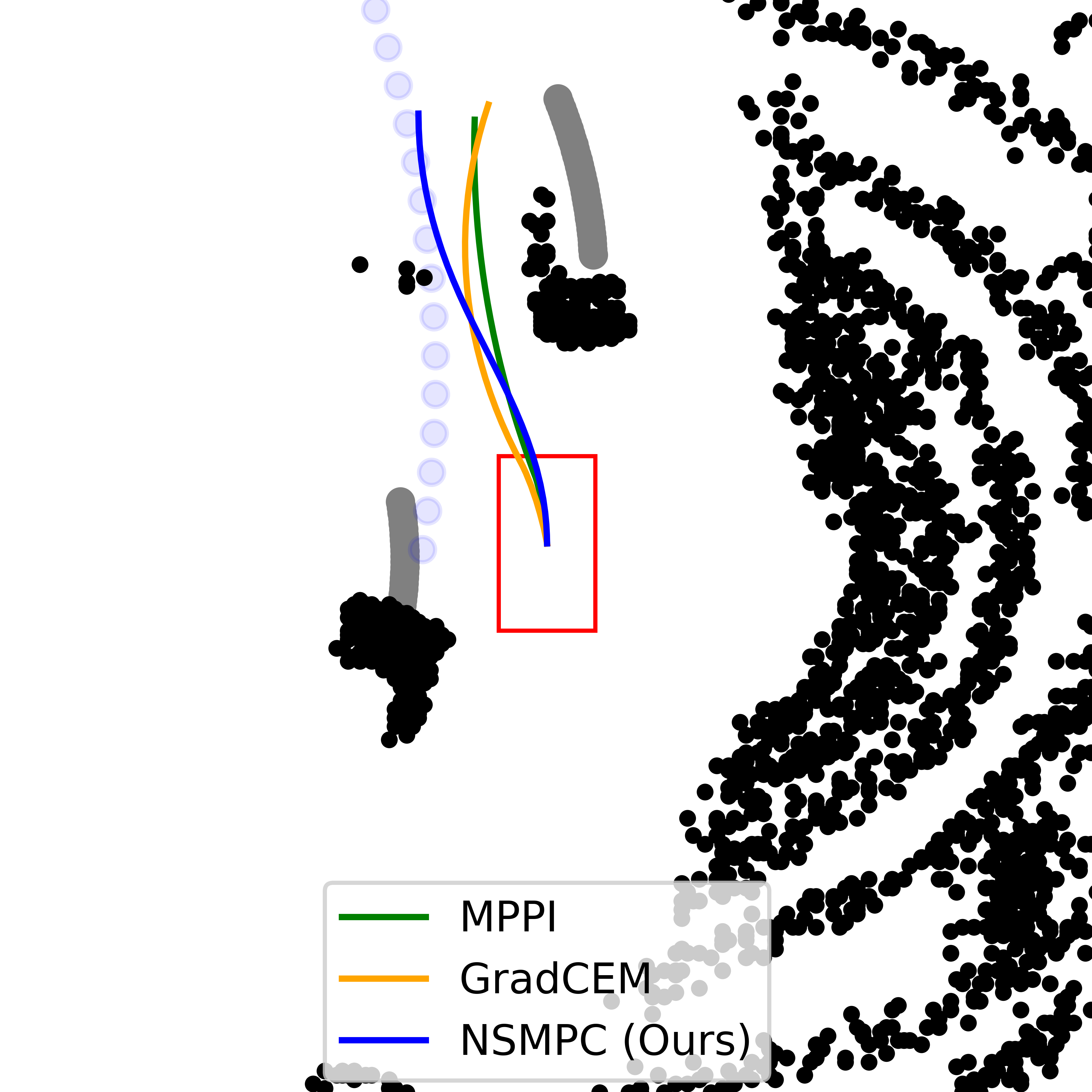}
      \end{subfigure}
      \begin{subfigure}{0.44\linewidth}
        \includegraphics[width=\linewidth]{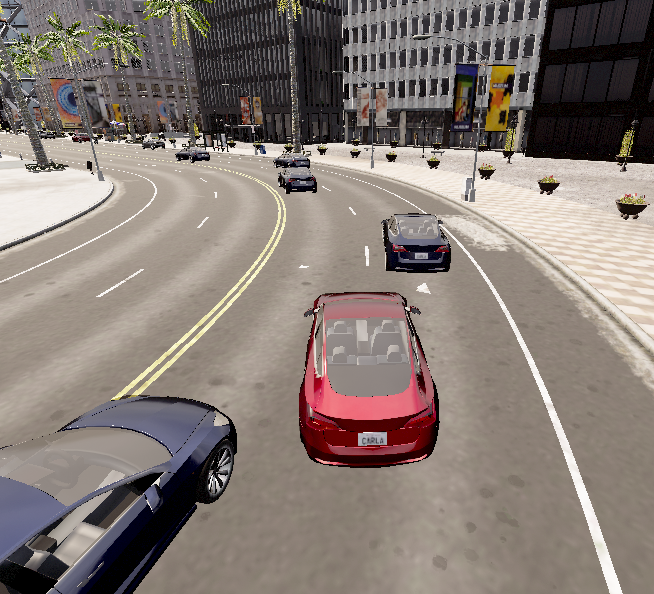}
      \end{subfigure}
      \caption{Obstacle avoidance} \label{fig:qa_st_lane_scene}
    \end{subfigure}
    
    \caption{We show a qualitative comparison of the trajectories generated by NSMPC (our method), MPPI and GradCEM in three different driving conditions: Straight empty road, Curved empty road and a road with Obstacles. The grey line denotes the future trajectory of the dynamic obstacles.} \label{fig:environments}
\end{figure}

In this section, we show a qualitative comparison between the trajectories generated by our NeuroSMPC (NSMPC) formulation and trajectories generated by Model Predictive Path Integral (MPPI) and gradient based Cross-Entropy Method (GradCEM) approaches. All the methods generate a short horizon ($30$ timesteps) control (velocity and angular-velocity) sequence. These controls are rolled out using the unicycle kinematics model to generate the trajectory. The NSMPC takes the past $5$ birds-eye view occupancy grid maps and the global path as input. For both the MPPI and GradCEM approach, the occupied cells in the occupancy grid map are considered as obstacles. In both the approaches the distribution is updated iteratively to get the mean controls. In Fig. \ref{fig:environments} we show the trajectory generated by our method compared to the trajectories generate by MPPI and GradCEM. We observe that the trajectory generated from our NSMPC formulation is qualitatively similar to the trajectories generated from the iterative approaches in an empty straight and curved road scenario and in a road with obstacles. 

\subsection{Quantitative Analysis}\label{quant_analusis}
\textbf{RoadSeg Network:} The RoadSeg Network, though only limited to the current operational area, allows for a much faster and accurate road segmentation with minimal GPU footprint. We compare our RoadSeg Network with two other baselines: RANSAC based plane segmentation and LSeg\cite{lseg} in terms of computation time and GPU memory utilization. From Table \ref{tab:roadseg_quant}, we can observe that we outperform other approaches by a significant margin. The lesser GPU footprint of RoadSeg allows us to run the entire NeuroSMPC pipeline on a single laptop in our AutoDP. 

\begin{table}[ht]
    \centering
    \begin{tabular}{|c|c|c|}
        \hline
        \textbf{Approach} & \textbf{GPU Memory Utilization (MB)} & \textbf{Inference Time (sec)} \\
        \hline
        \hline
        RoadSeg & 790 & 0.0015 \\
        \hline
        RANSAC & - & 0.08 \\
        \hline
        LSeg & 3492 & 0.018 \\
        \hline
        
    \end{tabular}
    \caption{RoadSeg Performance compared against RANSAC and LSeg. RoadSeg results in lesser GPU footprint and faster inference.}
    \label{tab:roadseg_quant}
\end{table}

\textbf{Neural guided Sampling based MPC:} We compare our method with two other sampling based MPC baselines: MPPI and GradCEM, during inference with respect to the number of iterations required to reach an optimal trajectory. In Table \ref{tab:iteration_quant}, we show that using our method, we achieve an optimal trajectory in single-shot, whereas using MPPI or GradCEM we need to iteratively update the mean to get an optimal trajectory. For MPPI we need to update the mean of the distribution for $5$ iterations whereas for GradCEM we need to update the mean for $3$ iterations. 

\begin{table}[]
    \centering
    \begin{tabular}{|c|c|c|}
        \hline
        \textbf{Approach} & \textbf{No. of iteration} \\
        \hline
        \hline
        NSMPC (Ours) & 0 \\
        \hline
        MPPI & 5 \\
        \hline
        GradCEM & 3 \\
        \hline
        
    \end{tabular}
    \caption{Comparison of SMPC (our method) with MPPI and GradCEM in terms of no. of update iterations required to get an optimal trajectory.}
    \label{tab:iteration_quant}
\end{table}

\subsection{Sim2Real}

One of the key challenges of a deep learning based system is to ensure that a model trained on a dataset collected from a simulator can also run effectively in the real world. While it is very easy to generate large volumes of data from a simulator, it is very difficult to accurately model real world physical phenomenons like friction, impact and uncertainty in a simulator. This results in the dataset from simulator not representing the real physical world accurately. We were able to overcome this issue by using a simplified form of data. The occupancy grid map from the simulator lidar is very similar to the occupancy grip map from the real world Lidar. This is because the lidar configuration in the simulator is the same as our AutoDP. The only thing missing in the simulator is the sensor noise which we overcame by perturbing the lidar points by a small amount randomly. After training the model on synthetic data collected from a simulator, we were able to run the model on our AutoDP. Fig. \ref{fig:sim2real} shows the execution of the model on our AutoDP while avoiding an obstacle in a controlled on campus scene. While there were still some instances where the model's output was colliding with the obstacle, the sampling nature of our formulation ensures that we still get an obstacle avoiding trajectory from the distribution of trajectories sampled from the model's output as mean.


\subsection{Why Sample?}

\begin{figure}
    \begin{minipage}[c]{0.6\linewidth}
    \includegraphics[width=\linewidth]{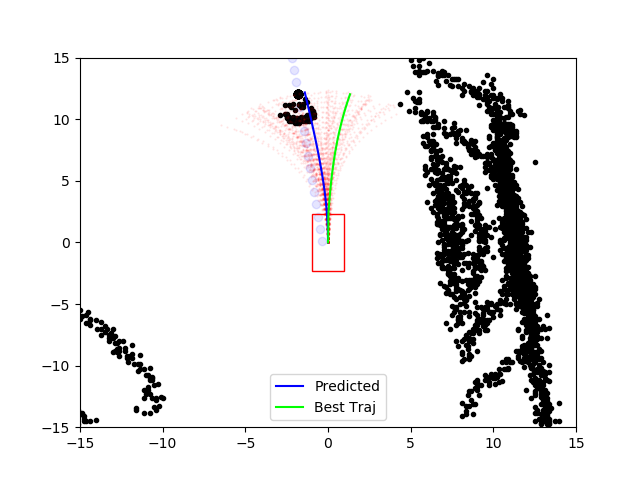}
  \end{minipage}\hfill
    \begin{minipage}[c]{0.4\linewidth}
    \caption{\textbf{Why sample?} Since the NN output may not always be collision free, we sample a distribution of trajectories (red) by using the NN output (blue) as the mean of a Gaussian distribution. The best trajectory (green) is then selected to be executed by the AutoDP.}
    \label{fig:why_sample}
    \end{minipage}
\end{figure}

The Neural Network's output is stochastic in nature and is not always guaranteed to avoid collisions with the obstacles. To ensure safety we need to make sure that the trajectory being executed by the AutoDP is always collision free. To solve this we sample a distribution of trajectories by using our model's output as a mean. The sampled trajectories are then scored using the cost functions mentioned in Section \ref{smpc}. The best trajectory is then selected from the sampled trajectories and executed by the AutoDP. Fig. \ref{fig:why_sample} shows an example scenario where the Neural Network output is colliding with the obstacle in front. The best trajectory (green) chosen from sampled trajectories is then executed by the AutoDP to avoid colliding with the obstacle.

\section{Implementation and Training}

The MPPI, GradCEM and our Neural guided Sampling based MPC has been implemented using the PyTorch \cite{NEURIPS2019_9015} library and trained on a system with a Intel Core i7-5930K (6 Core 12 threads) CPU and one NVIDIA RTX A4000 GPU with 16 GB VRAM and 64 GB RAM. We used the Lion optimizer \cite{chen2023symbolic} with a learning rate of 0.001 to train our network parameters. 


\section{Conclusion}

This paper proposed NeuroSMPC, a novel neural network guided Sampling Based Optimal Controller as an effective mechanism of interleaving single shot optimal inference with sampling based frameworks. NeuroSMPC overcomes the entailment of iterative resampling of sampling based optimal control frameworks by inferring single shot optimal trajectory rollouts and yet retains the advantage of sample diversity by sampling controls around the predicted rollouts. NeuroSMPC trajectory rollouts are similar to sampling based optimal control formulations \cite{bharadhwaj2020model, pmlr-v164-bhardwaj22a} yet come at a much faster clip thus making it suitable for real time settings like on-road autonmous driving. NeuroSMPC's spatio-temporal convolution architecture learns implicitly the dynamic nature of scenes without the need for explicit prediction of future states as it seamlessly avoids dynamic obstacles through an implicit understanding of their future evolution. NeuroSMPC has been tested in various CARLA scenes and a sim2real transfer on AutoDP (AutonomousDrivingPlatform) for on-road campus autonomous driving establishes its efficacy.

\section{Acknowledgements}

This work has partially been funded by Centre for Artificial Intelligence \& Robotics (CAIR).

\printbibliography

@inproceedings{zeng2019end,
  title={End-to-end interpretable neural motion planner},
  author={Zeng, Wenyuan and Luo, Wenjie and Suo, Simon and Sadat, Abbas and Yang, Bin and Casas, Sergio and Urtasun, Raquel},
  booktitle={Proceedings of the IEEE/CVF Conference on Computer Vision and Pattern Recognition},
  pages={8660--8669},
  year={2019}
}

@inproceedings{LSS, author = {Philion, Jonah and Fidler, Sanja}, title = {Lift, Splat, Shoot: Encoding Images from Arbitrary Camera Rigs by Implicitly Unprojecting to 3D}, year = {2020}, isbn = {978-3-030-58567-9}, publisher = {Springer-Verlag}, address = {Berlin, Heidelberg}, booktitle = {Computer Vision – ECCV 2020: 16th European Conference, Glasgow, UK, August 23–28, 2020, Proceedings, Part XIV}, pages = {194–210}, numpages = {17}, location = {Glasgow, United Kingdom} }

@InProceedings{pmlr-v164-bhardwaj22a,
  title = 	 {STORM: An Integrated Framework for Fast Joint-Space Model-Predictive Control for Reactive Manipulation},
  author =       {Bhardwaj, Mohak and Sundaralingam, Balakumar and Mousavian, Arsalan and Ratliff, Nathan D. and Fox, Dieter and Ramos, Fabio and Boots, Byron},
  booktitle = 	 {Proceedings of the 5th Conference on Robot Learning},
  pages = 	 {750--759},
  year = 	 {2022},
}

@inproceedings{bharadhwaj2020model,
  title={Model-predictive control via cross-entropy and gradient-based optimization},
  author={Bharadhwaj, Homanga and Xie, Kevin and Shkurti, Florian},
  booktitle={Learning for Dynamics and Control},
  pages={277--286},
  year={2020},
  organization={PMLR}
}

@incollection{NEURIPS2019_9015,
title = {PyTorch: An Imperative Style, High-Performance Deep Learning Library},
author = {Paszke, Adam and Gross, Sam and Massa, Francisco and Lerer, Adam and Bradbury, James and Chanan, Gregory and Killeen, Trevor and Lin, Zeming and Gimelshein, Natalia and Antiga, Luca and Desmaison, Alban and Kopf, Andreas and Yang, Edward and DeVito, Zachary and Raison, Martin and Tejani, Alykhan and Chilamkurthy, Sasank and Steiner, Benoit and Fang, Lu and Bai, Junjie and Chintala, Soumith},
booktitle = {Advances in Neural Information Processing Systems 32},
editor = {H. Wallach and H. Larochelle and A. Beygelzimer and F. d\textquotesingle Alch\'{e}-Buc and E. Fox and R. Garnett},
pages = {8024--8035},
year = {2019},
publisher = {Curran Associates, Inc.},
}

@inproceedings{sandler2018mobilenetv2,
  title={Mobilenetv2: Inverted residuals and linear bottlenecks},
  author={Sandler, Mark and Howard, Andrew and Zhu, Menglong and Zhmoginov, Andrey and Chen, Liang-Chieh},
  booktitle={Proceedings of the IEEE conference on computer vision and pattern recognition},
  pages={4510--4520},
  year={2018}
}

@inproceedings{Dosovitskiy17,
  title = { {CARLA}: {An} Open Urban Driving Simulator},
  author = {Alexey Dosovitskiy and German Ros and Felipe Codevilla and Antonio Lopez and Vladlen Koltun},
  booktitle = {Proceedings of the 1st Annual Conference on Robot Learning},
  pages = {1--16},
  year = {2017}
}

@misc{chen2023symbolic,
      title={Symbolic Discovery of Optimization Algorithms}, 
      author={Xiangning Chen and Chen Liang and Da Huang and Esteban Real and Kaiyuan Wang and Yao Liu and Hieu Pham and Xuanyi Dong and Thang Luong and Cho-Jui Hsieh and Yifeng Lu and Quoc V. Le},
      year={2023},
      archivePrefix={arXiv},
      primaryClass={cs.LG}
}

@inproceedings{shan2018lego,
  title={Lego-loam: Lightweight and ground-optimized lidar odometry and mapping on variable terrain},
  author={Shan, Tixiao and Englot, Brendan},
  booktitle={2018 IEEE/RSJ International Conference on Intelligent Robots and Systems (IROS)},
  pages={4758--4765},
  year={2018},
  organization={IEEE}
}

@misc{autoware,
  author={Autoware Foundation},
  title={Autoware: Open-source software for urban autonomous driving},
  year={2023},
}

@article{mildenhall2021nerf,
  title={Nerf: Representing scenes as neural radiance fields for view synthesis},
  author={Mildenhall, Ben and Srinivasan, Pratul P and Tancik, Matthew and Barron, Jonathan T and Ramamoorthi, Ravi and Ng, Ren},
  journal={Communications of the ACM},
  volume={65},
  number={1},
  pages={99--106},
  year={2021},
  publisher={ACM New York, NY, USA}
}

@inproceedings{CVAE,
  title={Multimodal Trajectory Predictions for Autonomous Driving
without a Detailed Prior Map},
  author={Kawasaki, Atsushi and Seki, Akihito},
  booktitle={2021 IEEE/CVF Winter Conference on Applications of Computer Vision},
  pages={3723--3732},
  year={2021},
  organization={IEEE}
}

@inproceedings{MP3,
  title={MP3: A Unified Model to Map, Perceive, Predict and Plan},
  author={Casas, Sergio and Abbas, Sadat and Urtasun, Raquel},
  booktitle={2021 IEEE Conference on Computer Vision and Pattern Recognition},
  pages={14403--14411},
  year={2021},
  organization={IEEE}
}

@inproceedings{INFER,
  title={INtermediate Representations for FuturE Prediction },
  author={Srikanth, Shashank and Ansari, Ahmed Junaid and Sharma, Sarthak and Jatavallabhula, Krishna Murthi and Krishna, Madhava},
  booktitle={2018 IEEE/RSJ International Conference on Intelligent Robots and Systems (IROS)},
  pages={4758--4765},
  year={2019},
  organization={IEEE}
}

@inproceedings{Desire,
  title={DESIRE: Distant Future Prediction in Dynamic Scenes with Interacting Agents },
  author={Lee, Namhon and Choi, Wongun and Paul, Vernazal and Torr, Philip and Chandraker, Manmohan},
  booktitle={2018 IEEE Conference on Computer Vision and Pattern Recognition (CVPR)},
  pages={336-344},
  year={2018},
  organization={IEEE}
}

@inproceedings{WACV-20,
  title={MonoLayout: Amodal Scene Layout from a Single Image},
  author={Mani, Kaustubh and Daga, Swapnil and Garg, Shubhika and Narasimhan, Sai Shankar and Jatavallabhula, Krishna Murthi and Krishna, Madhava},
  booktitle={2021 IEEE/CVF Winter Conference on Applications of Computer Vision},
  pages={3723--3732},
  year={2021},
  organization={IEEE}
}

@inproceedings{Kendall,
  title={Orthographic Feature Transform for Monocular 3d Object Detection},
  author={Roddick, Thomas and Kendall, Alex and Cippolla, Robert},
  booktitle={2019 British Machine Vision Conference},
  year={2019},
  organization={BMVA}
}

@inproceedings{CEM-Efficient,
  title={Sample-efficient Cross-Entropy Method
for Real-time Planning },
  author={Pinneri, Cristina and Sawany, Samburaj and Blaes, Sebastien and Stuckler, Jorh and Martius, George},
  booktitle={2020 4th Conference on Robot and Learning (CoRL)},
  year={2020},
  organization={CoRL}
}

@article{CEM,
  title={The Cross-Entropy Method: A Unified Approach to Combinatorial Optimization},
  author={Rubinstein, R.Y. and Kroese, D.P.},
  journal={Monte-Carlo Simulation, and Machine Learning, Springer-Verlag,},
  year={2004}
}

@misc{MPPI,
      title={Model Predictive Path Integral Control using Covariance Variable
Importance Sampling}, 
      author={Williams, Grady and Aldrich, Andrew and Theodorou, Evangelos},
      year={2015},
      archivePrefix={arXiv},
      primaryClass={cs.LG}
}

@article{lseg,
  title={Language-driven semantic segmentation},
  author={Li, Boyi and Weinberger, Kilian Q and Belongie, Serge and Koltun, Vladlen and Ranftl, Ren{\'e}},
  journal={arXiv preprint arXiv:2201.03546},
  year={2022}
}

@article{conceptfusion,
  title={ConceptFusion: Open-set Multimodal 3D Mapping},
  author={Jatavallabhula, Krishna Murthy and Kuwajerwala, Alihusein and Gu, Qiao and Omama, Mohd and Chen, Tao and Li, Shuang and Iyer, Ganesh and Saryazdi, Soroush and Keetha, Nikhil and Tewari, Ayush and others},
  journal={arXiv preprint arXiv:2302.07241},
  year={2023}
}

@inproceedings{werling2010optimal,
  title={Optimal trajectory generation for dynamic street scenarios in a frenet frame},
  author={Werling, Moritz and Ziegler, Julius and Kammel, S{\"o}ren and Thrun, Sebastian},
  booktitle={2010 IEEE International Conference on Robotics and Automation},
  pages={987--993},
  year={2010},
  organization={IEEE}
}

\end{document}